\pdfoutput=1

\documentclass[11pt]{article}

\usepackage[final]{acl}

\usepackage{times}
\usepackage{latexsym}

\usepackage[T1]{fontenc}

\usepackage[utf8]{inputenc}

\usepackage{microtype}

\usepackage{tabularx}

\usepackage{graphicx}
\graphicspath{ {./figures/} }

\usepackage{amsmath}
\usepackage{amssymb}
\usepackage{amsfonts}
\usepackage{amstext}
\usepackage{amsthm}
\usepackage{xcolor}

\title{RAAT: Relation-Augmented Attention Transformer for Relation Modeling in Document-Level Event Extraction}

\author{Yuan Liang\footnotemark[1], Zhuoxuan Jiang\footnotemark[1],  Di Yin and Bo Ren \\ Tencent Cloud, China \\
        \{ericyliang, alexzxjiang, endymecyyin, timren\}@tencent.com }

\begin{document}
\maketitle
\def\thefootnote{*}\footnotetext{These authors contributed equally to this work}
\begin{abstract}

In document-level event extraction (DEE) task, event arguments always scatter across sentences (across-sentence issue) and multiple events may lie in one document (multi-event issue). 
In this paper, we argue that the relation information of event arguments is of great significance for addressing the above two issues, and propose a new DEE framework which can model the relation dependencies, called Relation-augmented Document-level Event Extraction (ReDEE). More specifically, this framework features a novel and tailored transformer, named as Relation-augmented Attention Transformer (RAAT). RAAT is scalable to capture multi-scale and multi-amount argument relations. To further leverage relation information, we introduce a separate event relation prediction task and adopt multi-task learning method to explicitly enhance event extraction performance. Extensive experiments demonstrate the effectiveness of the proposed method, which can achieve state-of-the-art performance on two public datasets. Our code is available at \url{https://github.com/TencentYoutuResearch/RAAT}.
\end{abstract}

\section{Introduction}

Event extraction (EE) task aims to detect the event from texts and then extracts corresponding arguments as different roles, so as to provide a structural information for massive downstream applications, such as recommendation \citep{Li-Gao-re,Chun-Yi-CPMF}, knowledge graph construction \citep{Xindong-graph,Antoine-graph} and intelligent question answering \citep{Jordan-QA,Qingqing-QA}.

\begin{figure*}
    \centering
    \includegraphics[width=\textwidth]{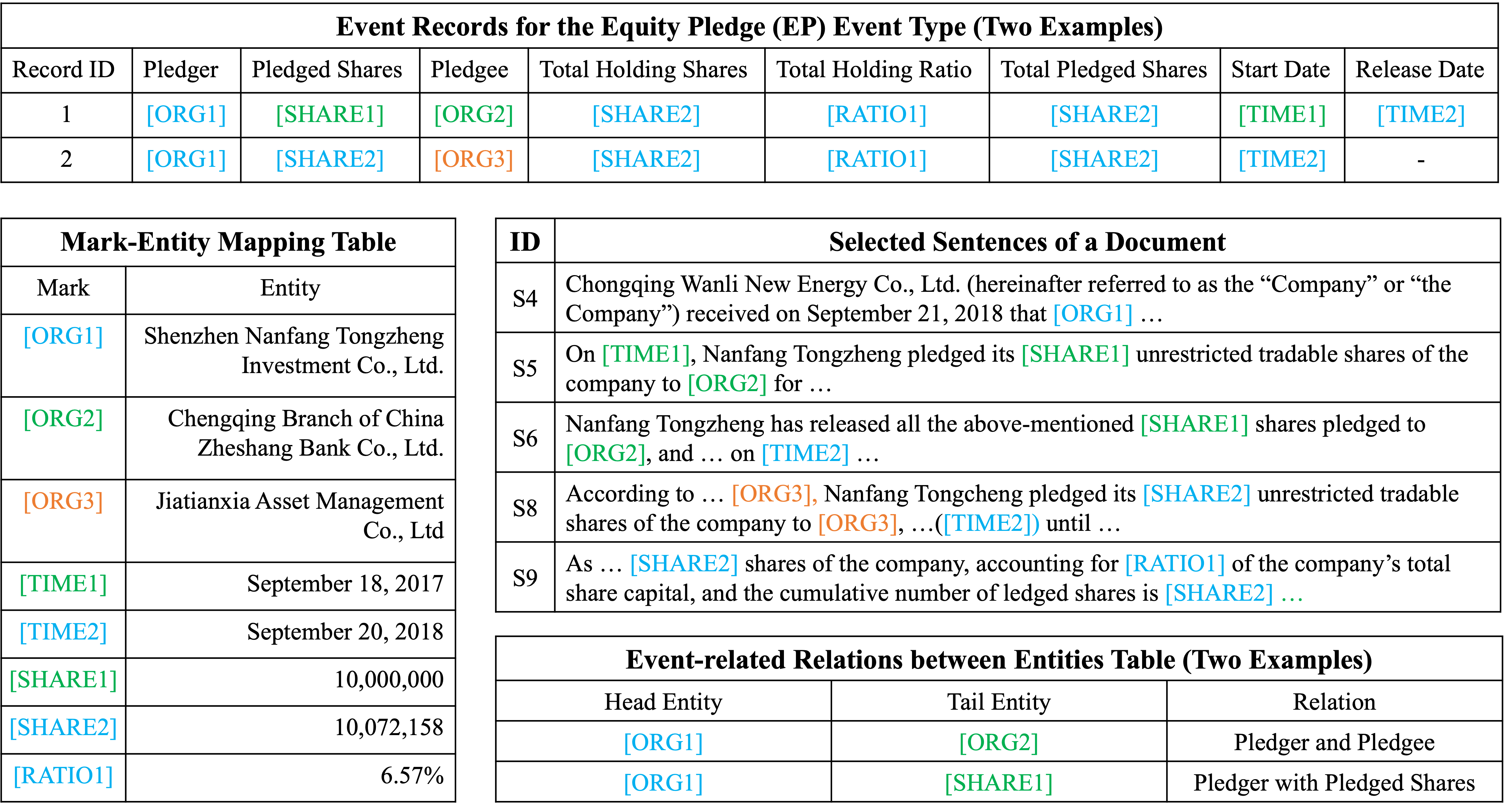}
    \caption{An example document for the event type of Equity Pledge, including selected sentences that are involved in multiple event records and where the event arguments scatter across sentences. We can observe that the relations between these entity mentions have intuitive patterns that could be leveraged to enhance the event extraction task. More information of entity color and complete event-related relations can be found in Appendix \ref{section:A.2}.}
    \label{fig:example}
\end{figure*}
Most of the previous methods focus on sentence-level event extraction (SEE) \citep{David-see,Shasha-see,Qi-see,Yubo-see,Thien-see,Yue-see,Lei-see,Haoran-see,Xinya-see,Fayuan-see,Giovanni-see,Yaojie-see}, extracting events from a single sentence. However, SEE is mostly inconsistent with actual situations. For example, event arguments may scatter across different sentences. As illustrated in Figure \ref{fig:example}, the event argument [ORG1] of event role \textit{Pledger} is mentioned in Sentence 4 and the corresponding argument [ORG2] of event role \textit{Pledgee} is in Sentence 5 and 6. We call this \textbf{across-sentence issue}. Another situation involves the \textbf{multi-event issue}, which means that multiple events may exist in the same document. As seen in the example in Figure~\ref{fig:example}, where two event records coincide, we should recognize that they may partially share common arguments.

Recently, document-level event extraction (DEE) attracts great attention from both academic and industrial communities, and is regarded as a promising direction to tackle the above issues \citep{DCFEE,Doc2EDAG,GIT,DE-PPN,PTPCG}. However, by our observation, we discover that the relations between event arguments have patterns which are an important indicator to guide the event extraction. This information is neglected by existing DEE methods. Intuitively, the relation information could build long-range relationship knowledge of event roles among multiple sentences, which could relieve the across-sentence issue. For multi-event issue, shared arguments within one document could be distinguished to different roles based on the different prior relation knowledge. As illustrated in Figure \ref{fig:example}, [ORG1] and [ORG2] have a prior relation pattern of \textit{Pledger} and \textit{Pledgee}, as well as [ORG1] and [SHARE1] for the relation pattern between \textit{Pledger} and its \textit{Pledged Shares}. 
Therefore, the relation information could increase the DEE accuracy if it is well modeled.

In this paper, we propose a novel DEE framework, called Relation-augmented Document-level Event Extraction (ReDEE), which is able to model the relation information between arguments by designing a tailored transformer structure. This structure can cover multi-scale and multi-amount relations and is general for different relation modeling situations. We name the structure as Relation-augmented Attention Transformer (RAAT). To fully leverage the relation information, we introduce a relation prediction task into the ReDEE framework and adopt multi-task learning method to optimize the event extraction task. We conduct extensive experiments on two public datasets. The results demonstrate the effectiveness of modeling the relation information, as well as our proposed framework and method.


In summary, our contributions are as follows:
\begin{itemize}
\item We propose a Relation-augmented Document-level Event Extraction (ReDEE) framework. It is the first time that relation information is implemented in the document-level event extraction field.

\item We design a novel Relation-augmented Attention Transformer (RAAT). This network is general to cover multi-scale and multi-amount relations in DEE. 

\item We conduct extensive experiments and the results demonstrate that our method outperform the baselines and achieve state-of-the-art performance by 1.6\% and 2.8\% F1 absolute increasing on two datasets respectively.
\end{itemize}

\section{Related Work}

\subsection{Sentence-level Event Extraction}

Previously, most of the related works focus on sentence-level event extraction. For example, a neural pipeline model is proposed to identify triggers first and then extracts roles and arguments \cite{Yubo-see}. Then a joint model is created to extract triggers and arguments simultaneously via multi-task learning \citep{Thien-see,Lei-see}. To utilize more knowledge, some studies propose to leverage document contexts \citep{Yubo-see1,Yue-see}, pre-trained language models \cite{Sen-see}, and explicit external knowledge \citep{Jian-see,Meihan-see}. However, these sentence-level models fail to extract multiple qualified events spanning across sentences, while document-level event extraction is a more common need in real-world scenarios. 

\subsection{Document-level Event Extraction}

Recently, DEE has attracted a great attention from both academic and industrial communities. At first, the event is identified from a central sentence and other arguments are extracted from neighboring sentences separately \cite{DCFEE}. Later, an innovative end-to-end model Doc2EDAG, is proposed \cite{Doc2EDAG}, which can generate event records via an entity-based directed acyclic graph to fulfill the document-level event extraction effectively. Based on Doc2EDAG, there are some variants appearing. For instance, GIT \cite{GIT} designs a heterogeneous graph interaction network to capture global interaction information among different sentences and entity mentions.  It also introduces a specific Tracker module to memorize the already extracted event arguments for assisting record generation during next iterations. DE-PPN \cite{DE-PPN} is a multi-granularity model that can generate event records via limiting the number of record queries. Not long ago, a pruned complete graph-based non-autoregressive model PTPCG was proposed to speedup the record decoding and get competitive overall evaluation results \cite{PTPCG}. In summary, although those existing works target for solving across-sentence and multi-event issues of the DEE task from various perspectives, to our best knowledge, we conduct a pioneer investigation on relation modeling towards this research field in this paper.

\subsection{Trigger-aware Event Extraction}
Previously a lot of works(\citep{Heng-see, Shasha-see, Qi-see, Yubo-see, Thien-see, Xiao-see}) deal with event extraction in two stages: firstly, trigger words are detected, which are usually nouns or verbs that clearly express event occurrences. And secondly, event arguments, the main attributes of events, are extracted by modeling relationships between triggers and themselves. In our work, we unify task as a whole to avoid error propagation between sub-tasks.


\section{Preliminaries}
Firstly, we clarify several key concepts in event extraction tasks. 1) \textit{\textbf{entity}}: a real world object, such as person, organization, location, etc.2) \textit{\textbf{entity mention}}: a text span in document referring to an entity object. 3) \textit{\textbf{event role}}: an attribute corresponding a pre-defined field in an event. 4) \textit{\textbf{event argument}}: an entity playing a specific event role. 5) \textit{\textbf{event record}}: a record expressing an event itself, including a series of event arguments.

In document-level event extraction task, one document can contain multiple event records, and an event record may miss a small set of event arguments. Further more, a entity can have multiple event mentions.

\section{Methodology}

\begin{figure*}
    \centering
    \includegraphics[width=\textwidth]{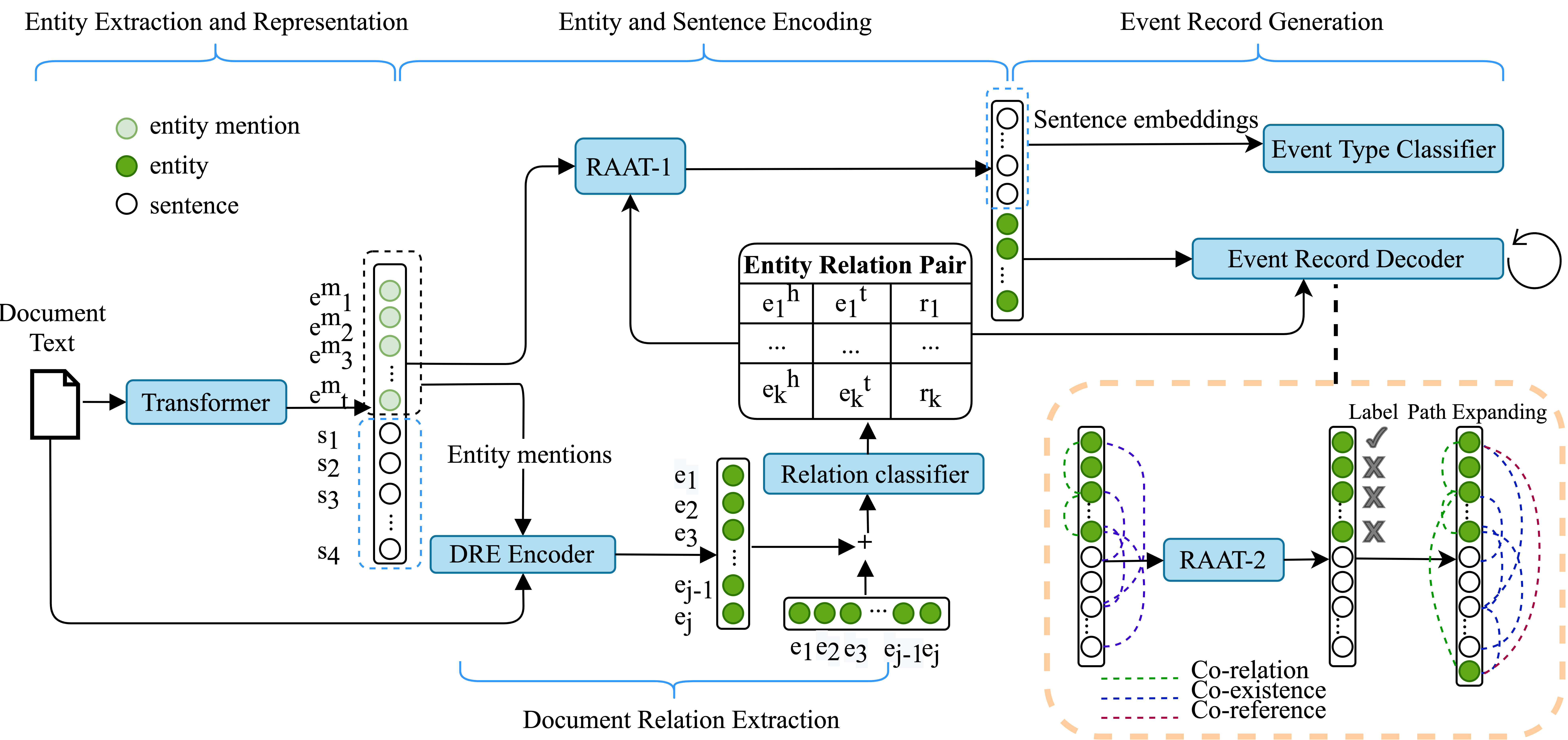}
    \caption{Overall of our proposed ReDEE framework.}
    \label{fig:model structure}
\end{figure*}

In this section, we introduce the proposed architecture first and then the key components in detail. 

\subsection{Architecture Overview}

End-to-end training methods for DEE usually involve a pipeline paradigm, including three sub-tasks: named entity recognition, event role prediction and event argument extraction. In this paper, we propose the Relation-augmented Document-level Event Extraction (ReDEE) framework coordinated with the paradigm. Our framework features leverage the relation dependency information in both encoding and decoding stages. Moreover, a relation prediction task is added into the framework to fully utilize the relation knowledge and enhance the event extraction task.

More specifically, shown in Figure~\ref{fig:model structure}, there are four key components in our ReDEE framework: Entity Extraction and Representation(EER), Document Relation Extraction(DRE), Entity and Sentence Encoding(ESE), and Event Record Generation(ERG). In the following, we would introduce the detailed definition of each component.


\subsection{Entity Extraction and Representation}

We treat the component of entity extraction as a sequence labeling task. Given a document $D$ with multiple sentences $\{s_1,s_2,...,s_i\}$, we use a native transformer encoder to represent the token sequence. Specifically, we use the BERT \cite{bert} encoder pre-trained in Roberta setting \cite{roberta}. Then we use the Conditional Random Field(CRF) \cite{CRF} to classify token representations into labels of named entities. We adopt the classical BIOSE sequence labeling scheme. The labels are predicted by the following calculation: $\hat{y}_{ne} = CRF(Trans(D))$. Then all the intermediate embeddings of extracted entity mentions and sentences are concatenate into a matrix $M_{ne+s}\in \mathbb{R}^{(j+i)\times{d_e}}$ by max-pooling operation on each sentence and entity mention span, where $j$ and $i$ are the numbers of entity mentions and sentences, and $d_e$ is the dimension of embeddings. The loss function for named entity recognition is denoted:
\begin{equation}
\small
    \mathcal{L}_{ne}= -\sum_{s_i \in D} logP(y_i|s_i)
\end{equation}
where $s_i$ denotes the $i^{th}$ sequence sentence in document, and $y_i$ is the corresponding ground truth label sequence. 

\subsection{Document Relation Extraction}

The DRE component takes the document text ($D$) and entities ($\{e_1,e_2,...,e_j\}$) extracted in the previous step as inputs, and outputs the relation pairs among entities, in a form of triples ($\{[e_1^h,e_1^t,r_1],[e_2^h,e_2^t,r_2],...,[e_k^h,e_k^t,r_k]\}$). [$e_k^h,e_k^t,r_k$] means the head entity, the tail entity and the relationship of the $k^{th}$ triple respectively. 

An important aspect is how to define and collect the relations from data. Here we assume that every two arguments in an event record can be connected by a relation. For example, \textit{Pledger} and \textit{Pledgee} in the \textit{EquityPledge} event could have a relation named as \textit{Pledge2Pledgee}, and the order of head and tail entities is determined by the pre-order of event arguments \cite{Doc2EDAG}. In this way, every event record with $n$ arguments could create $C_n^2$ relation samples. Note that this method to build relations is general to event extraction tasks from various domains, and the supervised relation information just comes from event record data itself, without any extra human labeling work. We do statistics for the relation types for ChiFinAnn dataset. Table~\ref{tab:stats snippet} shows a snippet of statistics and the full edition can be found in Appendix \ref{section:A.3}.

\begin{table}
\centering\small
\begin{tabular}{|l|l|l|l|}
\hline
\textbf{Relation Type} & \textbf{\#Train} & \textbf{\#Dev} & \textbf{\#Test} \\
\hline
Pledger2PledgedShares & 20002 & 2567 & 2299 \\
\hline
Pledger2Pledgee & 20002 & 2567 & 2299 \\
\hline
PledgedShares2Pledgee & 20002 & 2567 & 2299 \\
\hline
Start2EndDate & 19615 & 2239 & 1877 \\
\hline
Pledger2TotalHoldingShares & 18552 & 2412 & 2173 \\
\hline
\end{tabular}
\caption{The example relations with top 5 quantities in the ChiFinAnn dataset. The complete statistic can refer to the Appendix \ref{section:A.3}.}
\label{tab:stats snippet}
\end{table}

To predict the argument relations in this step, we adopt the structured self attention network~\cite{SSAN} which is the latest method for document-level relation extraction. However, different from previous work using multi-class binary cross-entropy loss, we use normal cross-entropy loss to predict only one label for each entity pair. The relation type is inferred by this function:
\begin{equation}
\small
    \hat{y}_{i,j}= argmax(e_i^TW_re_j)
\end{equation}
where $e_i, e_j \in \mathbb{R}^d$ denote entity embedding from encoder module of DRE and $d$ is the dimension of embeddings. $W_r\in \mathbb{R}^{d \times c \times d}$ denotes biaffine matrix trained by DRE task and $c$ is the total number of relations.
And the loss function for optimize the relation prediction task is denoted:
\begin{equation}
\small
    \mathcal{L}_{dre}= -\sum_{y_{i,j} \in Y} logP(y_{i,j}|D)
\end{equation}
where $y_{i, j}$ denotes ground truth label between the $i^{th}$ and $j^{th}$ entity, $D$ for document text and $Y$ for set of all relation pairs among entities.

\subsection{Entity and Sentence Encoding}

Now we have embeddings of entity mentions and sentences from EER component and a list of predicted triple relations from DRE component. Then this component encodes data mentioned above and output embeddings effectively integrated with relation information. In this subsection, we would introduce the method that translates triple relations to calculable matrices and the novel RAAT structure for encoding all the above data.

\begin{figure}
    \centering
    \includegraphics[height=8.5cm, width=7.5cm]{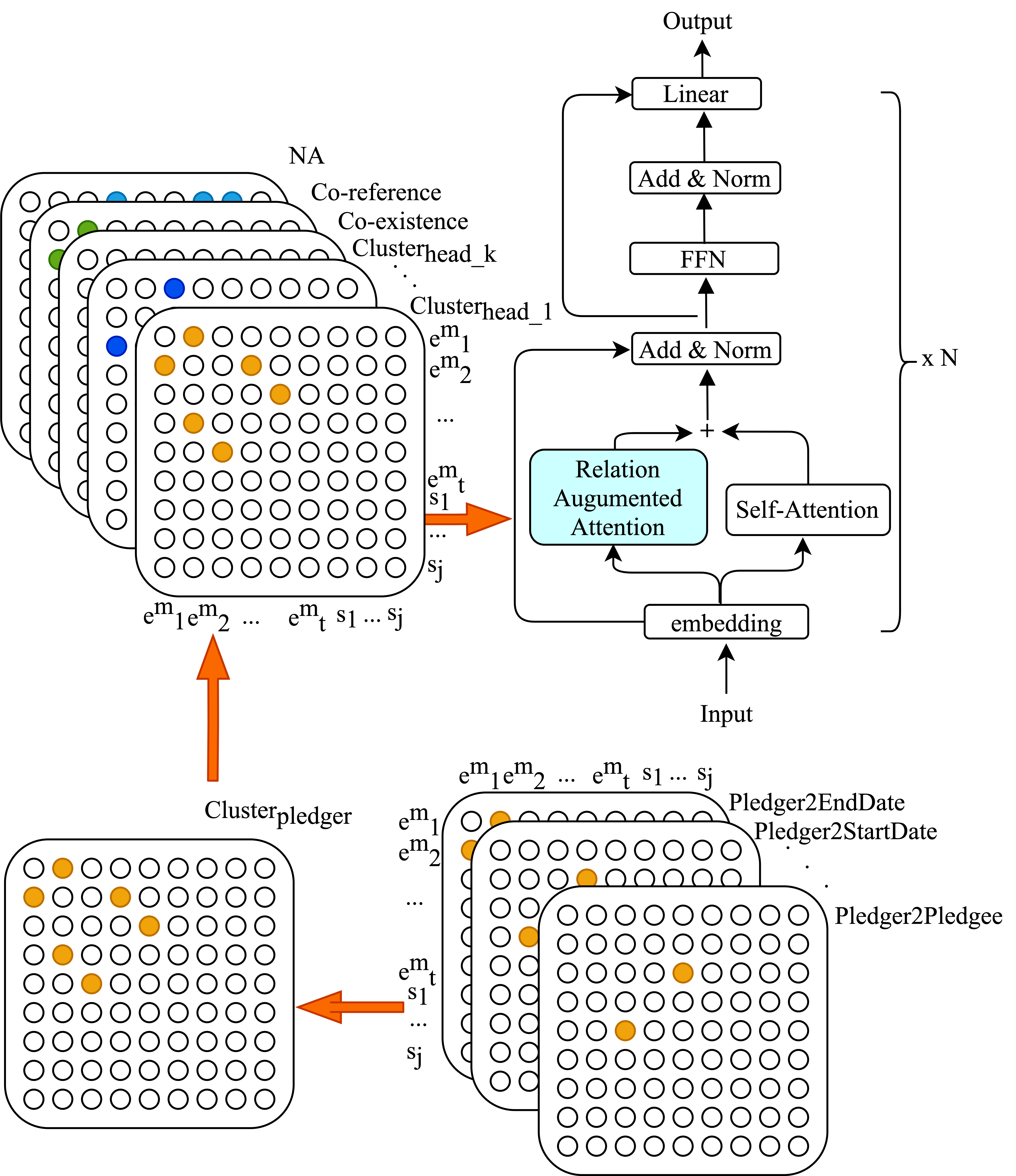}
    \caption{RAAT structure. Firstly each relation between entities and sentences are represented as matrices. Then the matrices are clustered by the head entities. At last the clustered matrices are integrated into the transformer structure for attention calculation.}
    \label{fig:raan structure}
\end{figure}

\subsubsection{Entity and Sentence Dependency}

First, we introduce a mechanism: entity and sentence dependency, which not only includes relation triples, but also describes links among sentences and entities beyond triples. 

\textit{Co-relation} and \textit{Co-reference} are defined to represent entity-entity dependency. For the former one, two entities have a \textit{Co-relation} dependency between them if they belong to a predicted relation triple. Entity pairs are considered having different \textit{Co-relation} if their involved triples have different relations. \textit{Co-reference} shows dependency between entity mentions pointing to same entities. That is, if an entity has several mentions existing across document, then each two of them has \textit{Co-reference} dependency. However, in the case that head and tail entities in relation triple are the same (i.e. \textit{StartDate} and \textit{EndDate} share same entities in some event records), then \textit{Co-relation} and \textit{Co-reference} are both held between them.

We use \textit{Co-existence} to describe dependency between entities and sentences where entity mentions come from. To be more specific, the entity mention together with its belonged sentence has \textit{Co-existence}. For remaining entity-entity and entity-sentence pairs without any dependency mentioned above, we uniformly treat them as \textit{NA} dependency. 

Table~\ref{tab:dependency system} shows the complete dependency mechanism. \textit{Co-relation} differs from \textit{NA}, \textit{Co-reference}, and \textit{Co-existence} in that it has several sub-types, with number equaling to that of relation types defined in document relation extraction task.
\begin{table}
    \centering\small
    \begin{tabular}{|c|m{7em}|m{7em}|}
    \hline
    & \textbf{sentence} & \textbf{entity} \\
    \hline
    \textbf{sentence} & NA & Co-existence/NA \\
    \hline
    \textbf{entity} & Co-existence/NA & Co-relation/Co-reference/NA \\
    \hline
    \end{tabular}
    \caption{All types of dependency among sentences and entities}
    \label{tab:dependency system}
\end{table}

\subsubsection{RAAT}
In order to effectively encode entity and sentence dependencies, we design the RAAT which takes advantage of a calculable matrix representing dependencies and integrates it into attention computation. According to the architecture shown in Figure~\ref{fig:raan structure}, RAAT is inherited from native transformer but has a distinct attention computation module which is made up of two parts: self-attention and relation-augmented attention computation.

Given a document shown as $D = \{s_1, s_2, ... s_j\}$, all entity mentions in this document as $E^m = \{e^m_{1}, e^m_{2}, ..., e^m_t\}$, where $e^m_i$ denotes entity mentions with the superscript $m$ denotes mention, and the subscript $i$ denotes index,
and a list of triples $\{[e_1^h,e_1^t,r_1],[e_2^h,e_2^t,r_2],...,[e_k^h,e_k^t,r_k]\}$, we build a matrix $T\in \mathbb{R}^{c \times (t + j) \times (t + j)}$ where $c$ for the number of dependencies, and $t$ and $j$ for the number of sentences and entity mentions respectively. $T$ is comprised of c matrices with same dimensions $R\in \mathbb{R}^{(t + j) \times (t + j)}$, and each $R$ represents one type of dependency $r \in \{Co-relation_k, Co-reference, Co-existence, NA\}, k = 1, 2, ... N$, $N$ as the number of relation types. For element within $T$, $t_{k, i, j}$ represent the dependency between $node_i$ and $node_j$. Specifically, $t_{k, i, j} = 1$ if they have the $k^{th}$ dependency, otherwise, $t_{k, i, j} = 0$. Here, $node_k \in \{e^m_{1}, e^m_{2}, ..., e^m_t, s_1, s_2, ... s_j\}$ can be either entity mention or sentence.

However, $T$ would be giant and sparse if we use the above strategy. To squeeze $T$ and decrease training parameters, we cluster \textit{Co-relation}  dependency based on the type of head entity in relation triple. For example, \textit{Pledger2Pledgee} and \textit{Pledger2PledgedShares} are clustered as one Co-relation dependency, and two matrice $R_a$ and $R_b$ corresponding to them are merged into one matrix. As a result, we finally get $T \in \mathbb{R}^{(3 + H) \times (t+j) \times (t + j)}$ where H denotes the number of head entity type in \textit{Co-relation}, and 3 covers \textit{NA}, \textit{Co-reference}, and \textit{Co-existence}.
Let $X \in \mathbb{R}^{(t+j) \times d}$ as input embeddings of attention module, $W_{rq}, W_{rk}, W_q, W_k, W_v \in \mathbb{R}^{d \times d}$, $M \in \mathbb{R}^{(3 + H) \times d \times d}$ as weight matrices,  we compute relation-augmented attention in the following steps:
\begin{equation}
\small
    Q_r=XW_{rq}, K_r=XW_{rk}
\end{equation}
\begin{equation}
\small
    S_a = \frac{\sum_{i=1}^{3+H} Q_rM[i, :, :]K^T_r \cdot T[i,:,:]}{\sqrt{d}} + bias_i
\end{equation}
where $S_a$ denotes score matrix of relation-augmented attention, $\cdot$ denotes element-wise multiplication. We compute self attention score and combine it with $S_a$ in the following way:
\begin{equation}
\small
    Q=XW_q, K=XW_k, W_v = XW_v
\end{equation}
\begin{equation}
\small
    S_b = \frac{QK^T}{\sqrt{d}}
\end{equation}
\begin{equation}
\small
    O = (S_a + S_b)V
\end{equation}
where $O$ is the output of attention module. Similar to the structure of native transformer, RAAT has multiple identical blocks stacking up layer by layer. Furthermore, $T$ is extensive since the number of \textit{Co-relation} can be selected. RAAT can be adaptive to the change of input length, which is equivalent to the total number of sentences and entity mentions.

\subsection{Event Record Generation}

With the outputs from previous component, the embeddings of entities and sentences, this ERG component actually includes two sub-modules: event type classifier and event record decoder.

\subsubsection{Event Type Classifier}

Given the embeddings of sentences, we adopt several binary classifiers on every event type to predict whether the corresponding event is identified or not. If there is any classifier identifying an event type, the following event record decoder would be activated to iteratively generate every argument for the corresponding event type. The loss function to optimize this classifier is as the following:
\begin{equation}
\small
    \mathcal{L}_{pred}= - \sum_i log(P(y_i|S))
\end{equation}
where $y_i$ denotes the label of the $i^{th}$ event type, $y_i = 1$ if there exists event record with event type i, otherwise, $y_i = 0$. $S$ denotes input embeddings of sentences.


\subsubsection{Event Record Decoder}

To iteratively generate every argument for a specific event type, we refer to the entity-based directed acyclic graph (EDAG) method~\cite{Doc2EDAG}. EDAG is a sequence of iterations with the length equaling to number of roles for certain event type. The objective of each iteration is to predict event argument of certain event role.
Inputs of each iteration are come up with entities and sentences embeddings. And the predicted arguments of outputs will be a part of inputs for next iteration. However, different from EDAG, we substitute its vanilla transformer part with our proposed RAAT structure (i.e. RAAT-2 as shown in Figure \ref{fig:model structure}). More specifically, the EDAG method uses a memory structure to record extracted arguments and adds role type representation to predict current-iteration arguments. However, this procedure hardly captures dependency between entities both in memory and argument candidates and sentences. In our method, RAAT structure can connect entities in memory and candidate arguments via relation triples extracted by the DRE component, and it can construct a structure to represent dependencies. In detail, before predicting event argument for current iteration, Matrix $T$ is constructed in the way shown above so that dependency is integrated into attention computation. After extracting the argument, it is added into memory, meanwhile, a new $T$ is generated to adapt next iteration prediction.

Therefore, the RAAT can strengthen the relation signal for attention computation. The RAAT-2 has the same structure with RAAT-1 but independent parameters. The formal definition of loss function for event recorder decoder is:
\begin{equation}
\small
    \mathcal{L}_a= -\sum_{v \in V_D}\sum_elog(P(y_e|(v, s)))
\end{equation}
where $V_D$ denotes node set in event records graph, $v$ denotes extracted event arguments of event record by far, $s$ denotes embedding of sentences and event argument candidates, and $y_e$ denotes label of argument candidate $e$ in current step. $y_e = 1$ means $e$ is the ground truth argument corresponding to current step event role, otherwise, $y_e = 0$.

\subsection{Model Training}

To train the above four components, we leverage the multi-task learning method \cite{Ronan-multi-task} and integrate the four corresponding loss functions together as the following: 
\begin{equation}
\small
    \mathcal{L} = \lambda_1\mathcal{L}_{ne} + \lambda_2\mathcal{L}_{dre} + \lambda_3\mathcal{L}_{pred} + \lambda_4\mathcal{L}_{a}
\end{equation}
where the $\lambda_i$ pre-set to balance the weight among the four components. 

\section{Experiments}

In this section, we report the experimental results to prove the effectiveness of our proposed method. In summary, the experiments could answer the following three questions:
\begin{itemize}
    \item To what degree does the ReDEE model outperform the baseline DEE methods?
    \item How well does ReDEE overcome across-sentence and multi-event issues?
    \item In what level does the each key component of ReDEE contribute to the final performance?
\end{itemize}


\begin{table*}
\centering
\resizebox{16cm}{1.8cm}{
\begin{tabular}{c|ccc|ccc|ccc|ccc|ccc|ccc}
\hline
\hline
\textbf{Model} & \multicolumn{3}{c|}{EF} & \multicolumn{3}{c|}{ER} & \multicolumn{3}{c|}{EU} & \multicolumn{3}{c|}{EO} & \multicolumn{3}{c|}{EP} & \multicolumn{3}{c}{Avg} \\ 
& P. & R. & F1.
& P. & R. & F1.
& P. & R. & F1.
& P. & R. & F1.
& P. & R. & F1.
& P. & R. & F1. \\
\hline
DCFEE-S$\dag$ & 61.1 & 37.8 & 46.7 & 84.5 & 86.0 & 80.0 & 60.8 & 39.0 & 47.5 & 46.9 & 46.5 & 46.7 & 64.2 & 49.8 & 56.1 & 67.7 & 54.4 & 60.3 \\
DCFEE-M$\dag$ & 44.6 & 40.9 & 42.7 & 75.2 & 71.5 & 73.3 & 51.4 & 41.4 & 45.8 & 42.8 & 46.7 & 44.6 & 55.3 & 52.4 & 53.8 & 58.1 & 55.2 & 56.6 \\
Greedy-Dec$\dag$ & 78.5 & 45.6 & 57.7 & 83.9 & 75.3 & 79.4 & 69.0 & 40.7 & 51.2 & 64.8 & 40.6 & 50.0 & 82.1 & 40.4 & 54.2 & 80.4 & 49.1 & 61.0 \\
Doc2EDAG$\dag$ & 78.7 & 64.7 & 71.0 & 90.0 & 86.8 & 88.4 & 80.4 & 61.6 & 69.8 & 77.2 & 70.1 & 73.5 & 76.7 & 73.0 & 74.8 & 80.3 & 75.0 & 77.5 \\
GIT$\dag$ & \textbf{78.9} & 68.5 & 73.4 & \textbf{92.3} & 89.2 & \textbf{90.8} & \textbf{83.9} & 66.6 & 74.3 & 80.7 & 72.3 & 76.3 & 78.6 & 76.9 & 77.7 & 82.3 & 78.4 & 80.3 \\
DE-PPN$\spadesuit$ & 78.2 & 69.4 & 73.5 & 89.3 & 85.6 & 87.4 & 69.7 & \textbf{79.9} & 74.4 & 81.0 & 71.3 & 75.8 & \textbf{83.8} & 73.7 & 78.4 & - & - & - \\
PTPCG$\clubsuit$ & - & - & - & - & - & - & - & - & - & - & - & - & - & - & - & \textbf{88.2} & 69.1 & 79.4 \\
\hline  
ReDEE(ours) & 78.0 & \textbf{70.6} & \textbf{74.1} & 91.1 & \textbf{90.3} & 90.7 & 82.5 & 69.2 & \textbf{75.3} & \textbf{83.7} & \textbf{73.1} & \textbf{78.1} & 81.7 & \textbf{78.6} & \textbf{80.1} & 84.0 & \textbf{79.9} & \textbf{81.9} \\		
\hline
\hline
\end{tabular}
}
\caption{Comparison of event extraction between baselines and our ReDEE model on the ChiFinAnn dataset. The missing parts are caused by the inaccessibility of baseline codes. $\dag$: results from \cite{GIT}; $\spadesuit$: results from  \cite{DE-PPN}; $\clubsuit$: results from  \cite{PTPCG}.}
\label{tab:chifinann1}
\end{table*}

\begin{table}
\centering\small
\resizebox{7.5cm}{1.2cm}{
\begin{tabular}{c|ccc|ccc}
\hline
\hline
\textbf{Model} & \multicolumn{3}{c|}{Dev} & \multicolumn{3}{c}{Online test} \\ 
& P. & R. & F1. & P. & R. & F1. \\
\hline
Doc2EDAG$\clubsuit$ & 73.7 & 59.8 & 66.0 & 67.1 & 51.3 & 58.1 \\
GIT$\clubsuit$ & 75.4 & 61.4 & 67.7 & \textbf{70.3} & 46.0 & 55.6 \\
PTPCG$\clubsuit$ & 71.0 & 61.7 & 66.0 & 66.7 & 54.6 & 60.0 \\
\hline
ReDEE(ours) & \textbf{77.0} & \textbf{72.0} & \textbf{74.4} & 69.2 & \textbf{57.4} & \textbf{62.8}\\
\hline
\hline
\end{tabular}
}
\caption{Comparison of event extraction between baselines and our ReDEE model on the DuEE-fin dataset. $\clubsuit$: results from  \cite{PTPCG}.}
\label{tab:dueefin1}
\end{table}

\subsection{Datasets}

DEE is a relatively new task and there are only a few datasets published. In our experiments we adopt two public Chinese datasets, i.e. \textbf{ChiFinAnn} \cite{Doc2EDAG} and \textbf{DuEE-fin} \cite{DuEE-fin}. 

ChiFinAnn includes 32,040 documents with 5 types of events, involving in equity-related activities for the financial domain. Statistics show that about 30\% of the documents contain multiple event records. We randomly split the dataset into train/dev/test sets in the ratio of 8/1/1. Readers can refer to the original paper for details.

DuEE-fin is also from the financial domain with around 11,900 documents in total. The dataset is downloaded from an online competition website\footnote{https://aistudio.baidu.com/aistudio/competition/detail/46}. Since there is no ground truth publicly available for the test set, we can only submit our extracted results to the website as a black-box online evaluation. Compared to ChiFinAnn, there are two differences. The DuEE-fin dataset has 13 different event types and its test set includes a large size of document samples that do not have any event records, which both make it more complicated. We get the distribution information of the dataset from Appendix \ref{section:A.1}.

\subsection{Baselines and Metrics}
Five different baseline models are taken into consideration: 1) \textbf{DCFEE} \cite{DCFEE}, the first model proposed to solve DEE task. 2) \textbf{Doc2EDAG} \cite{Doc2EDAG}, proposed an end-to-end model which transforms DEE as directly filling event tables with entity-based path expending. 3) \textbf{DE-PPN} \cite{DE-PPN}, a pipeline model firstly introducing the non-autoregressive mechanism. 4) \textbf{GIT} \cite{GIT}, a model using heterogeneous graph interaction network as encoder and maintaining a global tracker during the decoding process. 5) \textbf{PTPCG} \cite{PTPCG}, a light-weighted and latest DEE model. 

For evaluation metrics, we use precision, recall, and F1 score at the entity argument level for fair comparison with baselines. The overall "Avg" in the result tables denotes the micro average value of precision, recall, and F1 score. We conduct several offline evaluations for ChiFinAnn, but only an online test for DuEE-fin. 

\subsection{Settings}

In our implementation, for text processing, we consistently set the maximum sentence number and the maximum sentence length as 128 and 64 separately. We use BERT encoder in the EER component for fine-tuning and Roberta-chinese-wwm \cite{prtrainedModel} as the pre-trained model. Both RAAT-1 and RAAT-2 have four layers of identical blocks. More training details can be found in Appendix \ref{section:A.5}.
\begin{table}
\centering\small
\begin{tabular}{c|c|c|c|c}
\hline
\hline
\textbf{Model} & \textbf{I} & \textbf{II} & \textbf{III} & \textbf{IV} \\ 
\hline
DCFEE-S$\dag$ & 64.6 & 70.0 & 57.7 & 52.3 \\
DCFEE-M$\dag$ & 54.8 & 54.1 & 51.5 & 47.1 \\
Greedy-DEC$\dag$ & 67.4 & 68.0 & 60.8 & 50.2 \\
Doc2EDAG$\dag$ & 79.6 & 82.4 & 78.4 & 72.0 \\
GIT$\dag$ & 81.9 & 85.7 & 80.0 & 75.7 \\
\hline
ReDEE(ours) & \textbf{83.9} & \textbf{85.8} & \textbf{81.7} & \textbf{77.9} \\
\hline
\hline
\end{tabular}
\caption{F1 scores on four sets growing with average number of sentences involved in event records. $\dag$: results from \cite{GIT}.}
\label{tab:across-sentence}
\end{table}

\begin{table*}
\centering\small
\begin{tabular}{c|cc|cc|cc|cc|cc|ccc}
\hline
\hline
\textbf{Model} & \multicolumn{2}{c|}{EF} & \multicolumn{2}{c|}{ER} & \multicolumn{2}{c|}{EU} & \multicolumn{2}{c|}{EO} & \multicolumn{2}{c|}{EP} & \multicolumn{3}{c}{Avg} \\ 
& S. & M.
& S. & M.
& S. & M.
& S. & M.
& S. & M.
& S. & M. & S.\&M. \\
\hline
DCFEE-S$\dag$ & 55.7 & 38.1 & 83.0 & 55.5 & 52.3 & 41.4 & 49.2 & 43.6 & 62.4 & 52.2 & 69.0 & 50.3 & 60.3 \\
DCFEE-M$\dag$ & 45.3 & 40.5 & 76.1 & 50.6 & 48.3 & 43.1 & 45.7 & 43.3 & 58.1 & 51.2 & 63.2 & 49.4 & 56.6 \\
Greedy-Dec$\dag$ & 74.0 & 40.7 & 82.2 & 50.0 & 61.5 & 35.6 & 63.4 & 29.4 & 78.6 & 36.5 & 77.8 & 37.0 & 61.0 \\
Doc2EDAG$\dag$ & 79.7 & 63.3 & 90.4 & 70.7 & 74.7 & 63.3 & 76.1 & 70.2 & 84.3 & 69.3 & 81.0 & 67.4 & 77.5 \\
GIT$\dag$ & 81.9 & 65.9 & \textbf{93.0} & 71.7 & \textbf{82.0} & 64.1 & 80.9 & 70.6 & 85.0 & 73.5 & 87.6 & 72.3 & 80.3 \\
DE-PPN$\spadesuit$ & \textbf{82.1} & 63.5 & 89.1 & 70.5 & 79.7 & 66.7 & 80.6 & 69.6 & \textbf{88.0} & 73.2 & - & - & - \\
      
PTPCG$\clubsuit$ & - & - & - & - & - & - & - & - & - & - & \textbf{88.2} & 69.1 & 79.4 \\
\hline
ReDEE(ours) & 79.7 & \textbf{69.1} & 92.7 & \textbf{73.6} & 79.9 & \textbf{69.2} & \textbf{81.6} & \textbf{73.7} & 86.3 & \textbf{76.5} & 87.9 & \textbf{75.3} & \textbf{81.9} \\
\hline
\hline
\end{tabular}
\caption{Comparison of event extraction between singular (S.) and multiple (M.) event documents on the ChiFinAnn. $\dag$: results from \cite{GIT}; $\spadesuit$: results from  \cite{DE-PPN}; $\clubsuit$: results from  \cite{PTPCG}.}
\label{tab:multi-event}
\end{table*}

\begin{table}
\centering\small
\resizebox{7.5cm}{1.2cm}{
\begin{tabular}{l|ccc|ccc}
\hline
\hline
\textbf{Model} & \multicolumn{3}{c|}{ChiFinAnn} & \multicolumn{3}{c}{DuEE-fin}\\ 
& P. & R. & F1.
& P. & R. & F1.  \\
\hline
ReDEE & 84.0 & 79.9 & 81.9 & 69.2 & 57.4 & 62.8 \\
-RAAT-1  & +0.4 & -1.1 & -0.4 & +1.5 & -1.7 & -0.5 \\
-RAAT-2  & +1.3 & -2.4 & -0.7 & +0.8  & -3.2 & -1.7 \\
-RAAT-1\&2 & -3.1 & -0.1 & -1.5 & -1.3 & -5.1 & -3.7 \\
\hline
\hline
\end{tabular}
}
\caption{Ablation studies on ReDEE variants for RAAT.}
\label{tab:ablation1}
\end{table}

\subsection{Results and Analysis}

\paragraph{Overall Performance} Table \ref{tab:chifinann1} shows the comparison between baselines and our ReDEE model on the ChiFinAnn dataset. The ReDEE can achieve the state-of-the-art performance in terms of micro average recall and F1 scores on almost every type of events (i.e. EF, ER, EU, EO, EP), consistent with the Avg. results increased by 1.5\% and 1.6\% respectively. Our model also performs competitively well on precision results.

Table \ref{tab:dueefin1} shows the comparison results of our model with baselines on the developing set of DuEE-fin and its online testing. Seeing from former results, our model outperforms in a great leap by increasing 6.7\% on F1 score. For the online testing evaluation, our model has a distinct growth of 2.8\% on F1 score than the baselines. This experiment demonstrates our model could achieve a superior performance than existing methods.

\paragraph{Argument Scattering} The across-sentence issue widely exists in datasets. By our statistics, the training sets of ChiFinAnn and DuEE-fin have about 98.0\% and 98.9\% records that scatter across sentences respectively. To evaluate the performance of our model in different argument scattering degree, we compute the average number of sentences involved in records for each document and sort them in the increasing average number order. Then, all documents for testing are evenly divided into four sets, namely, I, II, III and IV, which means the I set is a cluster of documents that have the smallest average number of involved sentences while the IV set has the largest ones. According to table \ref{tab:across-sentence}, our model outperforms other baseline models in all settings, and meets the largest growth of 2.2\% F1 score in IV, the most challenging set of all. It indicates that our model is capable of capturing longer dependency of records across sentences via relation dependency modeling, thus alleviating the argument scattering challenge.
\paragraph{Single v.s. Multi Events} To illustrate how well our model performs in the multi-event aspect, we split the test set of ChiFinAnn into two parts: one for documents with single event record, and the other for documents including multiple events. Table \ref{tab:multi-event} shows the comparison results of all baselines and ReDEE. We find ReDEE performs much better in the multi-event scenario and outperforms baseline models dramatically in all five event types, improving ranging from 1.9\% to 3.2\% F1 scores. The results suggest that our relation modeling method is more effective to overcome the multi-event issue than existing baseline models.

\subsection{Ablation Study}

To probe the impact of RAAT structure for different components in ReDEE, we conduct ablation studies on whether to use RAAT or vanilla transformer.

In this experiment, we implement tests on three variants: 1) \textit{-RAAT-1}  substitutes the RAAT in the ESE component with vanilla transformer. 2) \textit{-RAAT-2} substitutes the RAAT in the event record generation module with vanilla transformer. 3) \textit{-RAAT-1\&2} substitutes the RAATs in both the above places with vanilla transformers, so that our model degrades to only import a relation extraction task via multi-task learning. 

The results in Table \ref{tab:ablation1} indicate that both two RAATs have positive influence on our model. Especially on ChiFinAnn, RAAT-2 makes more contribution than RAAT-1, with a decrease of 0.7\% versus 0.4\% in F1 scores once been substituted. After replacing both two RAATs, the value of relation extraction task becomes more weak and the model encounters a 1.5\% drop in F1 score. When it comes to DuEE-fin, a similar phenomenon can be observed that both the RAATs can contribute positively to our model.

\section{Conclusion}

In this paper, we investigate a challenging task of event extraction at document level, towards the across-sentence and multi-event issues. We propose to model the relation information between event arguments and design a novel framework ReDEE. This framework features a new RAAT structure which can incorporate the relation knowledge. The extensive experimental results can demonstrate the effectiveness of our proposed method which makes the state-of-the-art performance on two benchmark datasets. In the future, we will make more efforts to accelerate training and inference process.

\section*{Acknowledgements}
We thank the anonymous reviewers for their careful reading of our paper and their many insightful comments and suggestions. This work was supported  by Tencent Cloud and Tencent Youtu Lab. 

\bibliography{anthology,custom}
\bibliographystyle{acl_natbib}

\clearpage
\appendix
\section{Appendix}
In the appendix, we incorporate the following details that are omitted in the main body due to the space limit.

\subsection{Distribution of Event Type DuEE-fin}
\label{section:A.1}
Table~\ref{tab:duee_dist} shows the complete event type and corresponding distribution of DuEE-fin dataset. Overall, there are 13 event types in total with uneven distribution. Only train and development sets are shown since test set is not publicly available.

\begin{table}
\centering\small
\begin{tabular}{|c|c|c|}
\hline
\textbf{Event Type} & \textbf{\#Train.} & \textbf{\#Dev.} \\ 
\hline
ShareRedemption & 1309 & 243 \\
\hline
FinanceDeficit & 1062 & 163 \\
\hline
Pledge & 1027 & 160 \\
\hline
EnterpriseAcquisition & 934 & 142 \\
\hline
BidWin & 915 & 134 \\
\hline
ExecutiveChange & 901 & 134 \\
\hline
ShareholderHoldingDecrease & 876 & 147 \\
\hline
PledgeRelease & 728 & 118 \\
\hline
CorporateFinace & 535 & 72 \\
\hline
CompanyListing & 482 & 82 \\
\hline
ShareholderHoldingIncrease & 321 & 62 \\
\hline
CompanyBankruptcy & 236 & 44 \\
\hline
Admonition & 172 & 32 \\
\hline
\textbf{Total} & \textbf{9498} & \textbf{1533} \\
\hline
\end{tabular}
\caption{Distribution of Duee-fin dataset.}
\label{tab:duee_dist}
\end{table}


\subsection{Complete Relation Triples}
\label{section:A.2}
Table~\ref{appendix:crt} demonstrates the complete of relation triples of the document event extraction example shown in Figure~\ref{fig:example}. 

Entities in blue are involved in both two event records, while those in green and orange are exclusive to record 1 and 2 respectively. Heavy coupling of arguments among events increases the difficulty of multi-event issue.


\begin{table}\centering
    \includegraphics[height=11.5cm, width=7.5cm]{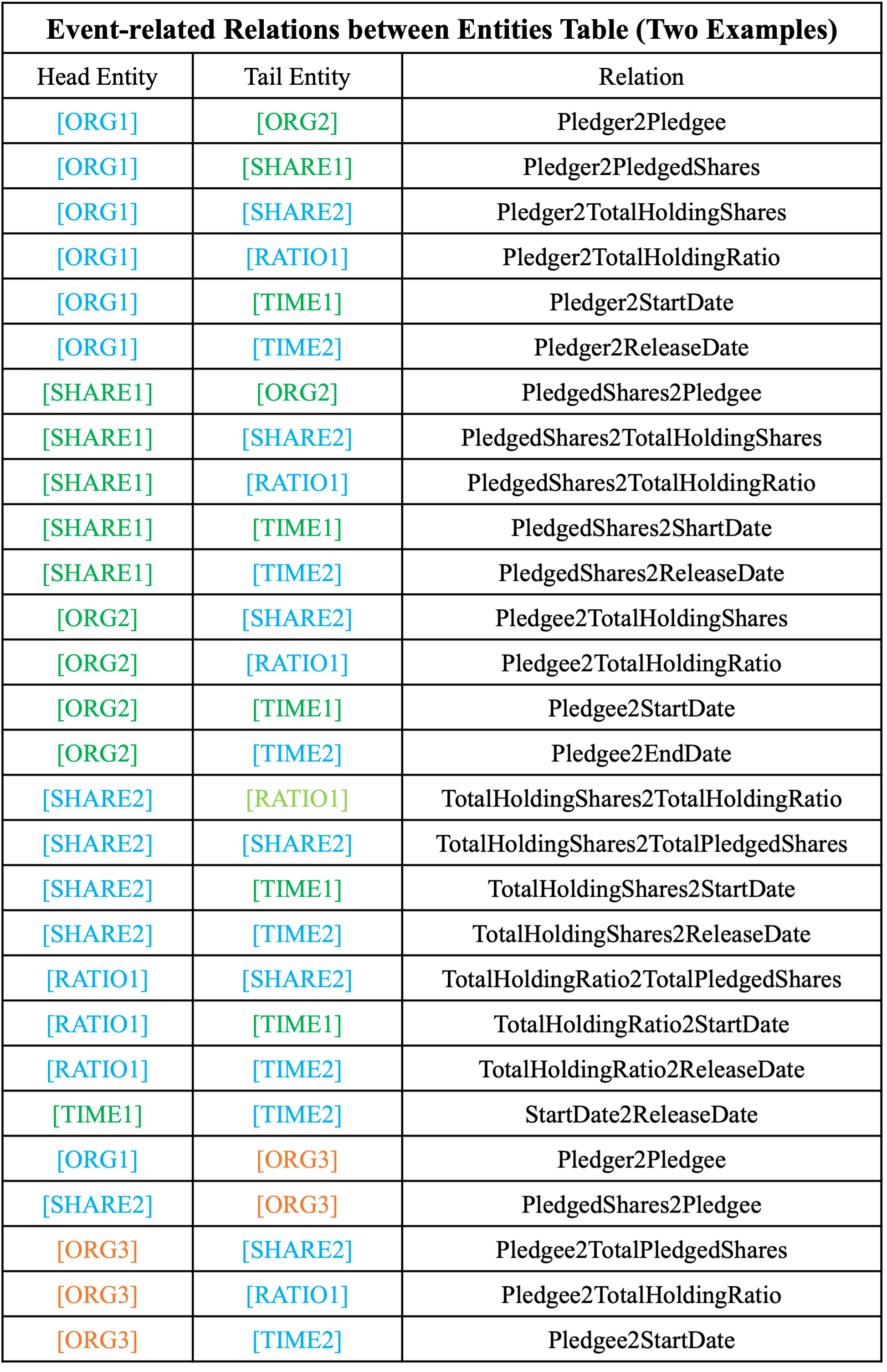}
    \caption{Complete relation triplets.}
    \label{appendix:crt}
\end{table}

\subsection{Relation Statistics for ChiFinAnn}
\label{section:A.3}
Table~\ref{appendix:re_stats} shows the relation statistics of ChiFinAnn dataset. There are 85 relation types in total, and train, development, and test sets have similar pattern in distribution.

\subsection{Case Study}
\label{section:A.4}

\begin{figure*}
    \centering
    \includegraphics[width=\textwidth]{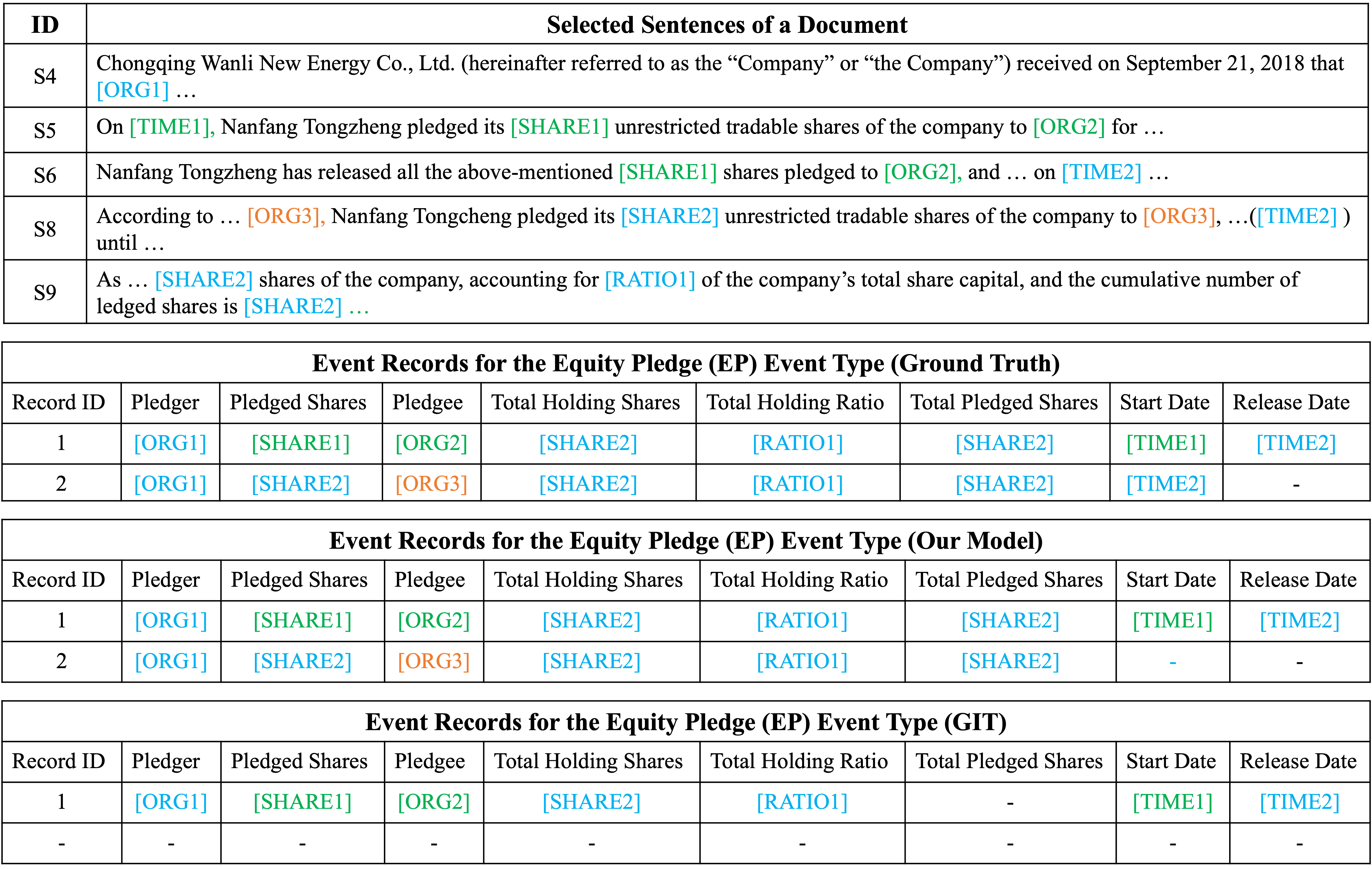}
    \caption{Case study.}
    \label{appendix:casestudy}
\end{figure*}

Figure~\ref{appendix:casestudy} shows the prediction results of our model and the best baseline model GIT on the example in Figure~\ref{fig:example}. Compared with the ground truth, our model correctly predicts all event arguments except one, while GIT only captures one event, with an argument missed. This example explicitly shows the superiority of our model in dealing with multi-events issue.

\subsection{More Training Settings}
\label{section:A.5}
For all native transformers and RAATs, the dimensions of hidden layers and feed-forward layers are set to 768 and 1,024 respectively. During training, we set the learning rate $lr=5e^{-5}$, batch size $b=64$. The four loss weights are set to $\lambda_1 = \lambda_3 = 0.05, \lambda2 = 1.0, \lambda_4=0.95$. We use 8 V100 GPUs and set gradient accumulation steps to 8. The train epoch are set to 100, and the best epoch are selected by the best validation score on development set for the evaluation of test set. And we use Adam to optimize the whole learning task.

\begin{table*}
    \centering
    \includegraphics[width=\textwidth]{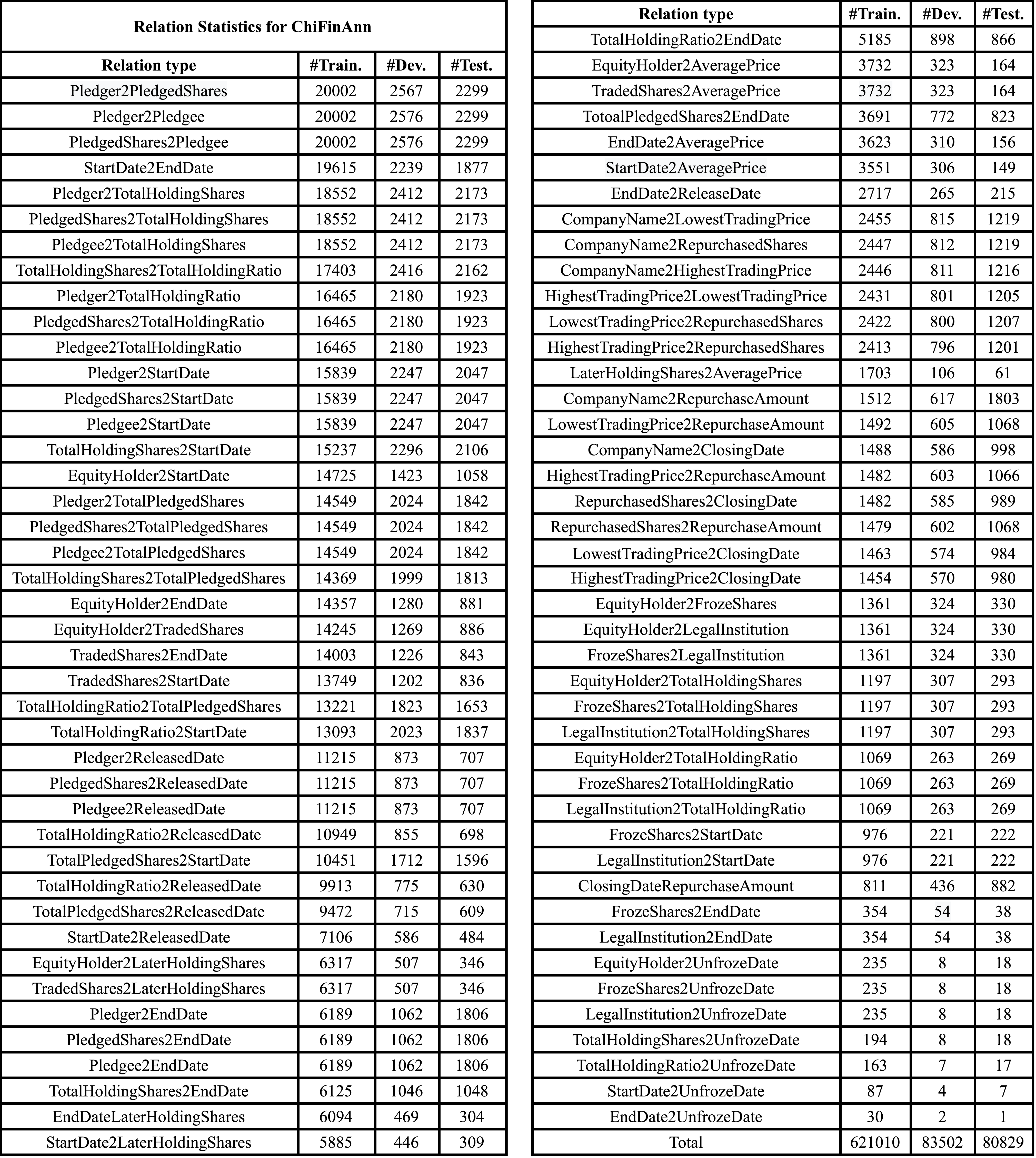}
    \caption{Relation statistics of ChiFinAnn dataset.}
    \label{appendix:re_stats}
\end{table*}

\end{document}